\documentclass[10pt,letterpaper]{article}
\usepackage[top=0.85in,left=.75in,footskip=0.75in,marginparwidth=2in]{geometry}

\usepackage[utf8]{inputenc}

\usepackage{cite}

\usepackage{nameref,hyperref}

\usepackage{microtype}
\DisableLigatures[f]{encoding = *, family = * }

\raggedright
\usepackage{changepage}

\usepackage[aboveskip=1pt,labelfont=bf,labelsep=period,singlelinecheck=off]{caption}

\makeatletter
\renewcommand{\@biblabel}[1]{\quad#1.}
\makeatother

\usepackage{lastpage,fancyhdr,graphicx}
\usepackage{epstopdf}
\fancyheadoffset[L]{2.25in}
\fancyfootoffset[L]{2.25in}

\usepackage{color}

\definecolor{Gray}{gray}{.25}

\usepackage{graphicx}

\usepackage{sidecap}

\usepackage{wrapfig}
\usepackage[pscoord]{eso-pic}
\usepackage[fulladjust]{marginnote}
\reversemarginpar

\usepackage[dvipsnames]{xcolor}

\usepackage{amsmath}
\usepackage{amsfonts}
\usepackage{multirow}

\newcommand{\x}{\mathbf{x}}
\newcommand{\R}{\mathbb{R}}
\newcommand{\X}{\mathbf{X}}
\renewcommand{\L}{\boldsymbol{\Lambda}}
\renewcommand{\O}{\mathbf{O}}
\renewcommand{\R}{\mathbb{R}}
\renewcommand{\S}{\mathbf{S}}

\graphicspath{{figures/}}

\begin{document}
\vspace*{0.35in}

\begin{flushleft}
{\Large
\textbf\newline{Deriving Visual Semantics from Spatial Context: An Adaptation of LSA and Word2Vec to generate Object and Scene Embeddings from Images}
}
\newline
\\
Matthias S. Treder\textsuperscript{1,*},
Juan Mayor-Torres\textsuperscript{2},
Christoph Teufel\textsuperscript{2}
\\
\bigskip
\bf{1} School of Computer Science, Cardiff University, United Kingdom
\\
\bf{2} Cardiff University Brain Research Imaging Centre (CUBRIC), School of Psychology, Cardiff University, United Kingdom
\\
\bigskip
* trederm@cardiff.ac.uk

\end{flushleft}

Embeddings are an important tool for the representation of word meaning. Their effectiveness rests on the distributional hypothesis: words that occur in the same context carry similar semantic information. Here, we adapt this approach to index visual semantics in images of scenes. To this end, we formulate a distributional hypothesis for objects and scenes: Scenes that contain the same objects (object context) are semantically related. Similarly, objects that appear in the same spatial context (within a scene or subregions of a scene) are semantically related. We develop two approaches for learning object and scene embeddings from annotated images. In the first approach, we adapt LSA and Word2vec's Skipgram and CBOW models to generate two sets of embeddings from object co-occurrences in whole images, one for objects and one for scenes. The representational space spanned by these embeddings suggests that the distributional hypothesis holds for images. In an initial application of this approach, we show that our image-based embeddings improve scene classification models such as ResNet18 and VGG-11 (3.72\% improvement on Top5 accuracy, 4.56\% improvement on Top1 accuracy). In the second approach, rather than analyzing whole images of scenes, we focus on co-occurrences of objects within subregions of an image. We illustrate that this method yields a sensible hierarchical decomposition of a scene into collections of semantically related objects. 
Overall, these results suggest that object and scene embeddings from object co-occurrences and spatial context yield semantically meaningful representations as well as computational improvements for downstream applications such as scene classification.



\section{Introduction}

Categorizing visual scenes quickly and robustly is critical for navigating an environment, localizing targets, and deciding how to act in a given context, both for humans  \cite{Malcolm2016MakingScenes} and robots \cite{Rangel2016SceneLabeling,Liao2016UnderstandNetworks}. Consequently, understanding scene perception is an important research topic in both biological and machine vision. The majority of computational approaches emphasize the importance of global summary statistics \cite{Oliva2001ModelingEnvelope}, or high-level features  \cite{Simonyan2015VeryRecognition} in scene categorization. Human observers, however, experience a scene as being composed of multiple objects, and the deeper meaning of a scene is determined by the physical and semantic relationships between these objects as well as their relationship to the scene gist \cite{Malcolm2016MakingScenes}. 

In general, visual scenes such as a room, a beach, or a parking lot are well-defined spatial locations that typically contain a large number of items arranged according to semantic (and syntactic) regularities. Importantly, scenes not only contain objects, which are defined as spatially distinct entities that can be moved or manipulated (e.g., a bar of soap, a stone, or a statue); they also incorporate “stuff”, a term that refers to amorphous areas that do not have these properties (e.g., walls, floor, a river, or the sky) \cite{Zhou2017SceneDataset,Zhou2018Places:Recognition}. In all of our analyses, objects and stuff are treated identically, and are assumed to be constituent parts of scenes. For brevity, we will therefore use the label \textit{objects} as an umbrella term to refer to both proper objects and stuff.

The purpose of this paper is to introduce vector representations of objects and scene categories and show how they might be useful for scene analysis and understanding. Importantly, our approach is not dependent on a text corpus, and we do not use image captions or complex annotations. Instead, we only rely on scene labels, object labels and the spatial organization of objects within a scene to develop two approaches for learning object and scene embeddings from images. In the first approach, the embeddings are based on co-occurrences of objects within scenes. In the second approach, we zoom into images and define as local spatial context the co-occurrence of objects in a small spatial window. 

In order to be useful, we aim for these representations to exhibit a number of properties critical for scene analysis and understanding: first, the embeddings should capture object-object relationships. For instance, we would like the embeddings to reflect the fact that a bar of soap and a towel are semantically closely related because they tend to co-occur in bathrooms and are used in the same functional contexts. Second, we hope to derive object-scene relationships such as a bar of soap being more closely related to a bathroom than to a living room. Third, we would like the embeddings to encode scene-scene relationships such that bathroom and bedroom are more closely related to each other than either of them to a football field. Finally, we aim for representations that are reliable at different hierarchical levels and can decompose scenes into sub-regions of semantically related objects.

In order to generate these image-based object and scene embeddings, we build on word embeddings, a technique that has been successfully employed to represent semantic relationships in natural language \cite{Lai2016HowEmbedding}. The motivation for word embeddings rests on the distributional hypothesis: words that occur in the same context, for instance, within the same sentence, tend to carry a similar meaning. Based on this assumption, the statistics of co-occurrences of words can be used as a proxy for semantic relatedness. 

In analogy to this idea, one can formulate the \textit{distributional hypothesis for objects and scenes}: scenes that contain the same objects (object context) are semantically related. Similarly, objects that appear in the same spatial context (other objects within the scene or within subregions of it) are semantically related. Note that, in analogy to the relationships between words in word embeddings, the relationships between objects and scenes are not functionally defined (e.g., toilet paper and toothbrush have very different functions) but purely governed by spatial proximity. Yet, empirical evidence supports the hypothesis that spatial relations mirror semantic relations. For instance, when Convolutional Neural Networks (CNNs) are trained to classify scenes, object detectors emerge as an intermediate representation of the network, suggesting that objects are informative with respect to scene category \cite{Zhou2014ObjectCNNs,Zhou2018Places:Recognition}.

\subsection{Related work}

Embeddings have been useful in domains other than text data. For instance, they are used to represent genes in gene-expression data \cite{Du2019Gene2vec:Co-expression}, computer network log data  \cite{Zhuo2017NetworkEmbeddings}, and graph-based data such as molecules \cite{Liu2019N-GramMolecules}.

In computer vision, several studies attempt to use image information in order to enhance word embeddings learnt from text corpora. For instance, \cite{Mao2016TrainingImages} add a feature vector derived from a VGG-16 CNN to a neural network that learns word embeddings from the respective image captions. Instead of using fixed image vectors, \cite{Ailem2020AImages} use a model based on a Variational Autoencoder that learns a latent visual representation along with the word embedding. \cite{Harwath2016DeepImages} learn multi-modal embeddings by combining images with spoken rather than textual image captions. \cite{Kottur2016VisualScenes} use embeddings initialized with Word2vec to predict annotations of abstract scenes, yielding modified embeddings that might better encode visual semantic relationships. Using a GloVe-based embedding model, \cite{Gupta2019ViCo:Co-occurrences} train scene embeddings from co-occurrences of objects within images.

A number of papers use features derived from objects in a scene to enhance scene classification. \cite{Li2012ObjectsClassification} propose an object filter bank to derive a set of image descriptors that serves as input for simple off-the-shelf classifiers such as SVM. \cite{Liao2016UnderstandNetworks} use semantic information from an image segmentation model as a regularizer for the first layers of an AlexNet CNN. A study by \cite{Chen2019SceneEmbeddings} is closest to ours. The authors propose to train object and scene embeddings from co-ocurrences within images. They then refine the prediction of a scene classifier using the predicted scene embedding. 

Our approach, too, relies on co-occurrences of objects. However, it differs from \cite{Chen2019SceneEmbeddings} in several important ways and offers the following additional contributions: (i) whereas previous papers focused on downstream applications such as scene classification only, we also investigate whether embeddings are semantically meaningful. Second, (ii) we adapt and compare a variety of models (LSA, Skipgram, CBOW). And third, (iii) we extend a state-of-the-art scene classification architecture with LSA embeddings. Unlike previous approaches, we directly fuse image features with object-based embeddings in the last CNN layer. Finally, (iv) we go beyond a 'bag-of-objects' approach and model local spatial context by considering the spatial proximity of objects within images.

\subsection{Dataset}
We use the ADE20K dataset \cite{Zhou2017SceneDataset} to train the object and scene embeddings (version  ADE20K\_2016\_07\_26). Unlike many other visual datasets, ADE20K contains a dense image annotation with every pixel being labeled. Additionally, each image is assigned to a scene category (e.g., abbey, bathroom). In total, the dataset contains 3,148 objects labels, 872 scene labels, and 22,210 individual images. We remove unlabeled objects and scenes, and only consider object and scene categories that occur in at least five images. An image has to contain at least two different objects to be selected. After applying these selection criteria, 1,140 object categories and 19,290 images remain in the dataset. The least common scene 'stone circle' appears in five images whereas the most common scene 'street' appears 2,241 times. The least common object 'carriage' appears in five images whereas the most common object 'wall' appears in 11,559 images. Sampling strategies, similar to those used for text data, are used to deal with this imbalance.

\section*{Scene embeddings from object co-occurrences}\label{sec:scene_embeddings}

\begin{figure*}
\centering
\includegraphics[width=0.9\textwidth]{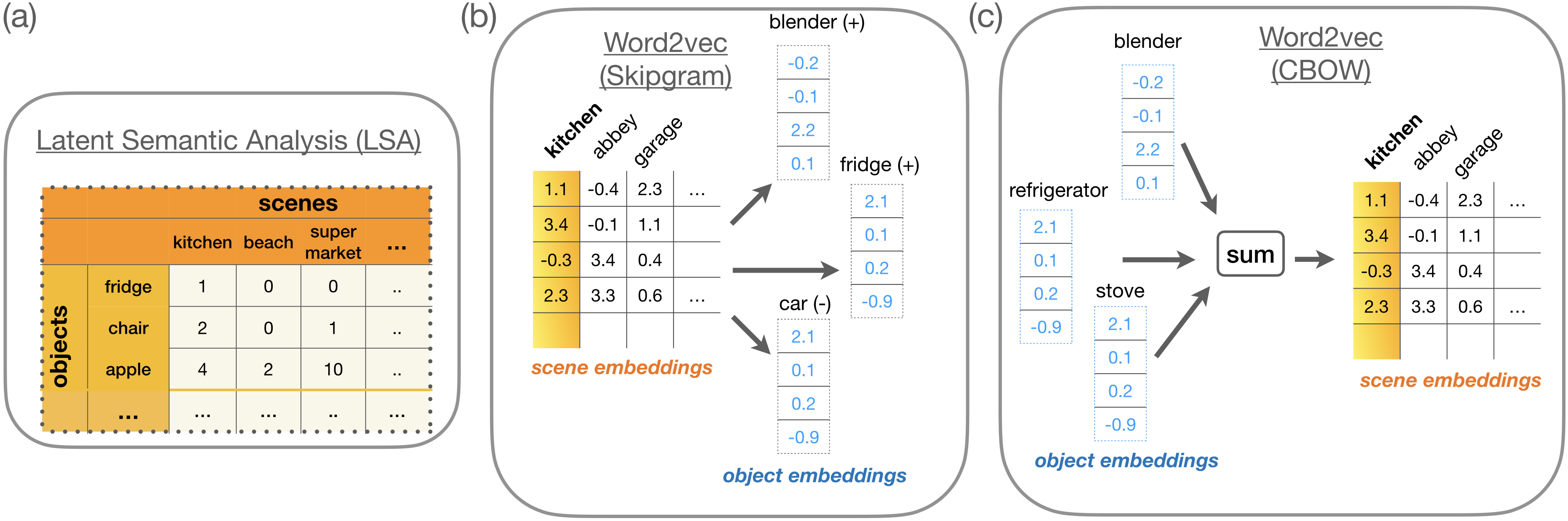}
\caption{Object and scene embeddings using LSA and Word2vec. (a) LSA operates on an object-scene occurrence matrix. (b) Single forward pass through a Skipgram model. A scene embedding (kitchen) is randomly selected. It is used to predict positive (blender, fridge) and negative (car) examples. (c) Forward pass through a CBOW model. Embeddings from a set of objects from a scene category are summed and used to predict the scene category.}
\label{fig:embedding_models}
\end{figure*}

We harness two word embedding algorithms, Latent semantic analysis (LSA)  \cite{Deerwester1990IndexingAnalysis} and Word2vec \cite{Mikolov2013EfficientSpace}, to learn two sets of embeddings, one for objects and one for scenes. For LSA, embeddings are obtained via matrix factorization of a scene-object co-occurrence matrix. For Word2vec, we either use a scene category to predict the objects it contains (Skipgram model) or use a set of objects to predict the category of the scene they appear in (CBOW). The resultant embeddings are used to test the distributional hypothesis for objects and scenes, and as feature vectors for a scene classification model.

\subsection{Latent semantic analysis (LSA)}

LSA models the relationship between words and a collection of documents they appear in  \cite{Deerwester1990IndexingAnalysis}. Its applications include document retrieval for search queries and topic modelling. It is based on the singular value decomposition (SVD) of a term-document matrix, wherein rows represent terms or words, columns represent documents, and each entry corresponds to the frequency of occurrence. In order to derive image-based embeddings using LSA, we replace the document-word matrix by a scene-object matrix  $\X\in\R^{n,m}$, where each row represents an object, each column represents a scene category, and the $(i,j)$-th entry is nonzero only if object $i$ appears in scene $j$. This is illustrated in Fig. \ref{fig:embedding_models}a. Using SVD, we can perform a low rank approximation of $\X$ as 

\begin{equation}
    \X\approx\O\L\S^\top.
\end{equation}

Here, the rows of $\O\in\R^{n\times d}$ act as object embeddings, $\S\in\R^{m\times d}$ as scene embeddings,  $\L\in\R^{d\times d}$ is a diagonal matrix of singular values, and $d$ is the embedding dimension which is selected by the user. To use this approach with ADE20k data, we first binarize the $1,140$-dimensional vector of object counts for each image. Subsequently, object vectors from images representing the same category are added together, yielding a $1,140\times 682$ matrix. Three different normalization approaches for the object counts are explored: \textit{LSA-norm} (divide each column by its total count), \textit{LSA-log} (log-transform the counts), and \textit{LSA-tfidf} (TF-IDF transform). TF-IDF is a popular normalization scheme  \cite{Beel2013ResearchSurvey} and is given by the product 

\[
\text{tfidf}(o, s) = \text{tf}(o,s)\cdot \text{idf}(o)
\]

where $\text{tf}(o,s)$ is the term frequency, i.e., the number of times an object $o$ occurs in a scene category $s$, and $\text{idf}(o) = m/|\sum_{i=1}^m \X(o,i)>0|$ is the inverse document frequency that down-weighs objects that occur in many scene categories. An advantage of LSA over methods such as Word2vec is that in addition to embeddings for items in the training vocabulary, it also provides a linear transform that can be applied to unseen images. For a vector of object counts $\x\in\R^n$ representing a test image, the respective embedding is given by 

\begin{equation}
\text{embed}(\x) = \O^\top\x.
\end{equation}

We will exploit this property in section \ref{sce:scene_classification} when we perform scene classification on test images. 

\subsection{Word2vec}\label{sec:word2vec}

Word2vec constructs word embeddings by relating target words to words with a context window, e.g., other words from the same sentence \cite{Mikolov2013EfficientSpace}. It comes in two flavors: In the Skipgram model, the conditional probability of a context word given the target word is maximized, whereas the CBOW model maximizes the probability of the target word given the sum of the embeddings of its context words. In typical text applications, input and output words come from the same domain, and two sets of embeddings (for inputs and outputs) are obtained. After training, the two sets are either averaged or one of them is selected to represent the embeddings.

Here, we consider an asymmetrical approach in which inputs and outputs stem from different domains, one representing objects and the other scenes. Therefore, one matrix contains the object embeddings, the other matrix contains the scene embeddings. For the Skipgram model (depicted in Fig. \ref{fig:embedding_models}b), the objective it to maximize the average log probability for an object $o$ given the scene $s$ it appears in,

\begin{equation}\label{eq:skipgram}
\sum_{s\in S}\sum_{o\in\text{context}(s)}\log p(o | s)
\end{equation}

where $S$ is the set of scene images and context(s) is defined as the set of objects present in the respective image. Although our dataset is comparably small, we consider the subsampling and negative sampling strategies from \cite{Mikolov_NIPS13} for the Skipgram model. To subsample scenes categories (inputs), the relative number of images from a given category is defined as its frequency $f(s)$. An image is rejected from the training data with a probability of $p(s) = 1-\sqrt{t/f(s)}$. Since the image corpus is relatively small, we set $t=0.005$ and the subsampling is repeated in every epoch. For an input image, five objects from the scene are chosen as positive outputs (objects are repeated for images with less than five objects). For negative sampling of objects, denote the frequency of an object as $f(o)$. We thus sample 20 negative objects with a probability of $p(o) = f(o)^{3/4}/\sum_{o'\in O}f(o)^{3/4}$, where $O$ is the set of all objects.

For the CBOW model, uniform sampling is used throughout. First, an object is randomly sampled. Then an image containing this object is randomly selected, and the remaining objects are sampled from this image. A context size of five objects is used, and the scene category corresponding to the image serves as target. Both models are implemented in Pytorch, using 100 epochs with an Adam optimizer and a learning rate of 0.01.

\subsection{Testing the distributional hypothesis for scenes}

\begin{figure}
\centering
\includegraphics[width=0.3\textwidth]{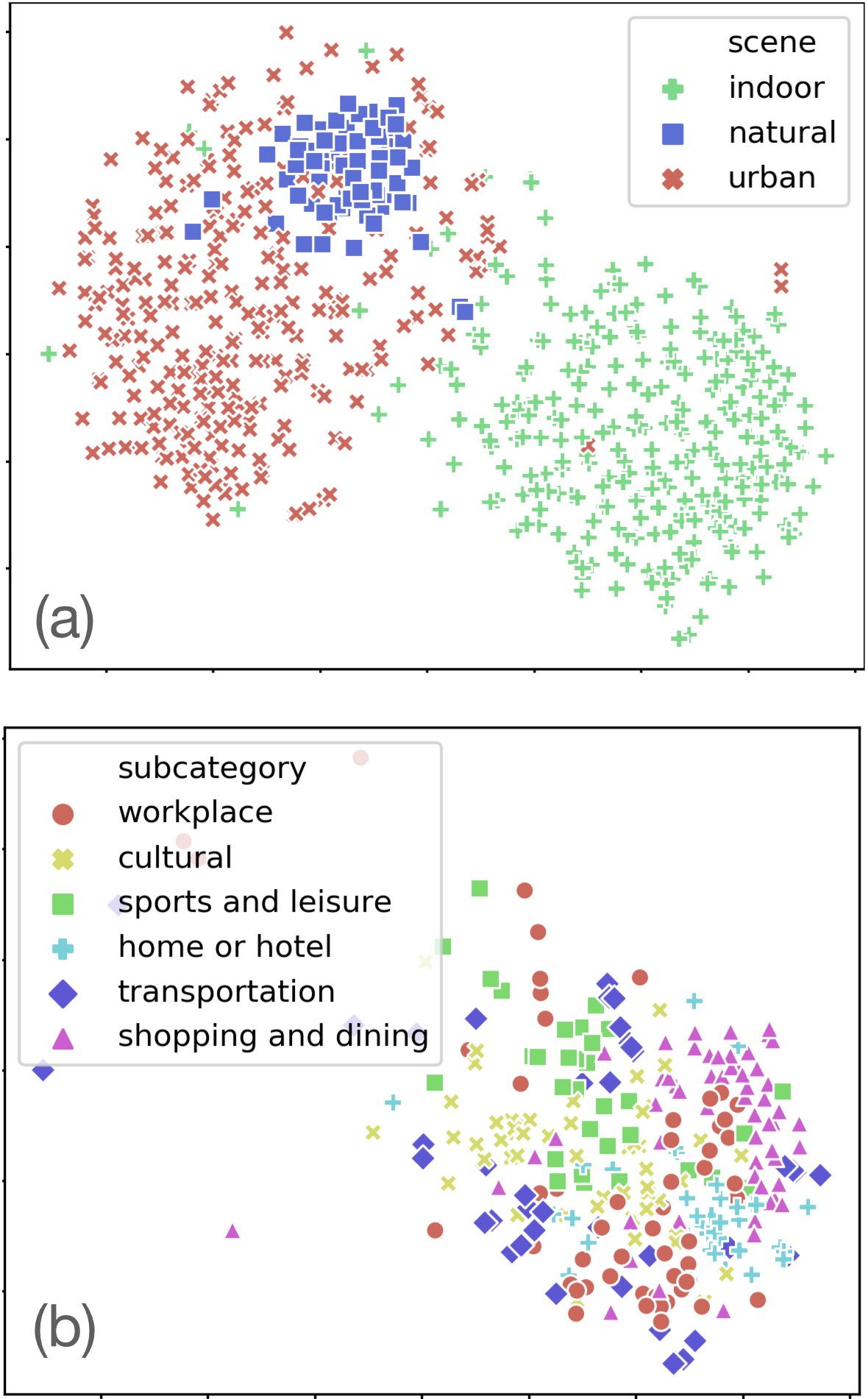}
\caption{t-SNE visualization of scene embeddings for the Skipgram model (d=100), using a cosine metric. (a) Supercategories. (b) Subcategories of 'indoor'.}
\label{fig:tsne}
\end{figure}

If the distributional hypothesis for scenes is correct and relevant semantic information is preserved in the embeddings, the embedding vectors should cluster according to semantic properties such as membership in supercategories. For instance, in the Places dataset \cite{Zhou2018Places:Recognition} scene categories can be indoor (e.g. bathroom), natural outdoor (e.g. beach), or urban outdoor (e.g. football field). In line with this, we find that scenes within these supercategories appear more similar to each other than to scenes from other supercategories, using a cosine distance metric. This is exemplarily depicted for the Skipgram model in Fig. \ref{fig:tsne}a. Indoor, urban and natural scenes form well-defined clusters. Urban and natural scenes are partially overlapping, which could be explained by the fact that they are both outdoor scenes and hence share a number of objects such as sky, sun, or mountain. 

As a more quantitative evaluation, the results for a Wilcoxon rank-sum test are reported in Table \ref{tab:wilcoxon} for all models and d=100. The comparison of the within-category vs between-categories cosine distances is highly significant and illustrates that scenes within a category are much more similar to each other than to scenes from other categories. This relationship holds for subcategories, too (Fig. \ref{fig:tsne}b). Results and visualizations for other embedding dimensions and all subcategories are provided in the supplementary material.

\begin{table}
\footnotesize
\begin{tabular}{ |p{1.5cm}|p{1.5cm}|p{1.5cm}|}
 \hline
 Method & Wilcoxon z & p-value\\
 \hline
{LSA-norm} & -16.33 & $<0.0001$\\
\hline
{LSA-tfidf} &  -14.72 & $<0.0001$\\
\hline
{LSA-log} & -15.62 & $<0.0001$\\
\hline
{Skipgram} & -261.18  & $<0.0001$\\
\hline
{CBOW} & -98.69 & $<0.0001$\\
\hline
\end{tabular}
\caption{Within vs between category Wilcoxon rank-sum test results (z-statistic and p-value) for an embedding dimension of d=100.}\label{tab:wilcoxon}
\end{table}

\subsection{Scene classification}\label{sce:scene_classification}

To evaluate the efficacy and relevance of the LSA embeddings, we compare scene classification performance of Residual Networks \cite{He2016DeepRecognition} (ResNets) and VGG networks \cite{Simonyan2015VeryRecognition} \emph{without LSA} features to the performance when the corresponding LSA features are appended to the second last fully connected layer. In a second type of evaluation, we add an \emph{object presence} vector rather than the LSA features. It is a 1200-dimensional binary array from the ADE20K Matlab API with an entry of 1 if an object is present in a particular scene and 0 otherwise. 

\subsubsection{ResNet18 and VGG11 training:} 

ResNet18 and VGG11 are trained in two classification analyses: first, (i) the training/validation split proposed by the ADE20K filenames, and second, (ii) a 5-fold cross-validation based on the combined training/validation data. For both analyses we use batches of 100 composed of randomly cropped 224$\times$224 px images. Each random crop is normalized using a uint8 normalization transform in Pytorch \cite{Paszke2019PyTorch:Library}. 

On the training/validation split we select the 506 scene categories contained in both sets. For the 5-fold cross-validation the number of classes depends on each fold (499-509). For ResNet18 we use an initial global learning rate of 0.1 with a linear weight decay of 0.0001 after each epoch. For VGG11, to prevent a divergence of the weight amplitudes we use an initial learning rate of 0.01 with a weight decay of $10^{-5}$ after each epoch. The models' weights and biases are initialized using the Glorot's initializer \cite{torres2017enhanced} and optimized using Adam. Models are trained for 100 epochs in all analyses. 

\subsubsection{Adding LSA features:} 
We add the LSA embeddings to two extra fully-connected (FC) ReLU layers added to the end of each model architecture as an additional classification subnetwork \cite{Zhang2016SPDA-CNN:Recognition}. 
The first FC layer has 4096 neurons. It receives inputs from both the 512 units of the last layer of ResNet18/VGG11 and the 300-dimensional vector of LSA embedding features. This feeds into a second FC ReLU layer with 4096 units which is followed by a final softmax layer. We use a fan-out/fan-in ratio greater than one on our proposed FC layers connection in order to optimize the layer-to-layer activation  \cite{Glorot2010UnderstandingNetworks}. 

LSA transforms are generated using the training exemplars in each analysis. The transform is then applied to the object annotations of the respective test images.
The weights and biases of the extra FC layers are updated using the same training parameters as in the baseline model. Therefore, we do not use ResNet18 and VGG11 with pre-trained weights but rather train the full model including the additional FC layers from scratch.

For consistency with the image batch generation we use Pytorch's \emph{SubsetRandomSampler} synchronized with the corresponding image batches to generate corresponding LSA training batches. The LSA embeddings are not randomly cropped. Rather, batch indices are permuted on each training iteration. This approach guarantees consistent and reliable Class-Activation maps as reported in the supplementary material \cite{Zhou2016LearningLocalization}.  

\subsubsection{Results:} Table \ref{tab:trainvalres} shows the scene classification results of ResNet18 and VGG11 pipelines on the training/validation split.  ResNet18 shows an average improvement of 3.72\% on Top1 and 4.56\% on Top5 accuracies when the LSA embeddings are included in the FC layer in comparison with the baseline \emph{without LSA} features. 

ResNet18 shows an average improvement of 3.72\% on Top1 and 4.56\% on Top5 accuracies when LSA embeddings are included as compared to the baseline without additional features. Compared to the baseline with the object presence vector, ResNet18 with LSA features shows an improvement of 2.99\% on Top1 and 3.77\% on Top5.
VGG11 does not show a significant improvement on any of the baselines. We hypothesize that the information from LSA embeddings is not properly absorbed by the VGG11 network due to the large amount of FC-layers connected at the end of the layer and a poor segmentation associated with a relatively large size average-pool layer compared to ResNet18 \cite{Shen2017DeepAnalysis}.

\begin{table*}[!ht]
\small
\begin{tabular}{|p{1.4cm}|c|p{1cm}|c|c|c|c|c|c|c|c|c|c|}
\hline
\textbf{Training /validation} & \multicolumn{2}{l|}{\textbf{Without LSA}} & \multicolumn{2}{l|}{\textbf{Object presence}}  & \multicolumn{2}{l|}{\textbf{LSA}} & \multicolumn{2}{l|}{\textbf{LSA categories}} & \multicolumn{2}{l|}{\textbf{LSA tf-idf}} & \multicolumn{2}{l|}{\textbf{LSA log}} \\ \cline{ 2 - 13}
\multicolumn{ 1}{|l|}{} & \textbf{Top1} & \textbf{Top5} & \textbf{Top1} & \textbf{Top5} & \textbf{Top1} & \textbf{Top5} & \textbf{Top1} & \textbf{Top5} & \textbf{Top1} & \textbf{Top5} & \textbf{Top1} & \textbf{Top5} \\ \cline{ 1 - 13}
\textbf{VGG11} & 44.56 & 65.34 & 42.88 & 65.88 & 44.65 & 64.41 & 45.34 & 65.11 & 45.07 & 66.12 & 44.79 & 64.33 \\ \cline{ 1- 13}
\textbf{ResNet18} & 49.01 & 71.06 & 50.77 & 71.65 & \textbf{53.47} & \textbf{75.45} & 51.06 & \textbf{73.77} & \textbf{52.37} & \textbf{75.88} & 50.35 & \textbf{73.94} \\ \hline
\end{tabular}
\caption{Training/validation results showing the Top1 and Top5 accuracies for the two baselines: (1) evaluating scene classification with the image data only or \emph{Without LSA}, and (2) using the object presence obtained from a binary \emph{Object Vector} included on the ADE20K API. The evaluations related to the LSA features are (1) using the \emph{LSA} normalized embeddings, (2) using the LSA embeddings obtained from \emph{combining the scene categories}, (3) using the LSA embeddings from the \emph{tf-idf} transform, and (4) the LSA embeddings after the \emph{log} transform. Values in bold are $p<0.001$ comparing the scores with the \emph{without LSA} baseline.}
\label{tab:trainvalres}
\end{table*}

\begin{table*}[!htb]
\small
\begin{tabular}{|p{1.4cm}|p{0.9cm}|p{0.9cm}|p{0.9cm}|p{0.9cm}|p{0.9cm}|p{0.9cm}|p{0.9cm}|p{0.9cm}|p{0.9cm}|p{0.9cm}|p{0.9cm}|p{0.9cm}|}
\hline
\textbf{5-fold cross-val} & \multicolumn{2}{l|}{\textbf{Without LSA}} & \multicolumn{2}{l|}{\textbf{Object presence}}  & \multicolumn{2}{l|}{\textbf{LSA}} & \multicolumn{2}{l|}{\textbf{LSA categories}} & \multicolumn{2}{l|}{\textbf{LSA tf-idf}} & \multicolumn{2}{l|}{\textbf{LSA log}} \\ \cline{ 2 - 13}
\multicolumn{ 1}{|l|}{} & \textbf{Top1} & \textbf{Top5} & \textbf{Top1} & \textbf{Top5} & \textbf{Top1} & \textbf{Top5} & \textbf{Top1} & \textbf{Top5} & \textbf{Top1} & \textbf{Top5} & \textbf{Top1} & \textbf{Top5} \\ \cline{ 1 - 13}
\textbf{VGG11} & 40.225$\pm$ 10.553 & 61.346$\pm$ 8.914 & 41.445$\pm$ 9.573 & 61.047$\pm$ 9.657 & 41.112$\pm$ 9.141 & 61.914$\pm$ 9.775 & 41.134$\pm$ 9.256 & 61.331$\pm$ 9.671 & 41.916$\pm$ 9.661 & 62.061$\pm$ 9.814 & 41.618$\pm$ 9.467 & 61.390$\pm$ 9.551 \\ \cline{ 1- 13}
\textbf{ResNet18} & 42.556$\pm$ 11.237 & 62.277$\pm$ 10.333 & 42.959$\pm$ 10.111 & 62.691$\pm$ 9.989 & 44.112$\pm$ 9.991  & 63.491$\pm$ 10.143 & 43.562$\pm$ 9.877 & 62.952$\pm$ 10.225 & 44.067$\pm$ 10.023 & 63.001$\pm$ 10.120 & 43.067$\pm$ 9.741 & 62.114$\pm$ 10.342 \\ \hline
\end{tabular}
\caption{5-fold cross-validation results showing the Top1 and Top5 accuracies. Standard deviations are calculated across the 5 folds.}
\label{tab:5fold}
\end{table*}

Table \ref{tab:5fold} shows the results for the 5-fold cross-validation modality in percentages. No significant improvements are found comparing any performance including LSA information with any performance baseline such as \emph{without LSA} and \emph{Object presence} information.
The lack of significant improvement observed on the 5-fold cross-validation might be attributable to a smaller size of the training set in comparison to the training/validation split. We hypothesize that the size of the training set is important for obtaining a significant performance improvement on deep feed-forward networks. However, the information of the LSA embeddings can be properly transferred using the extra FC layers as we report on the Class Activation Map (CAM) analysis. CAM and additional analyses are presented in the supplementary material.


\section{Object embeddings from spatial context}\label{sec:spatial_context}

\begin{figure*}
\centering
\includegraphics[width=0.9\textwidth]{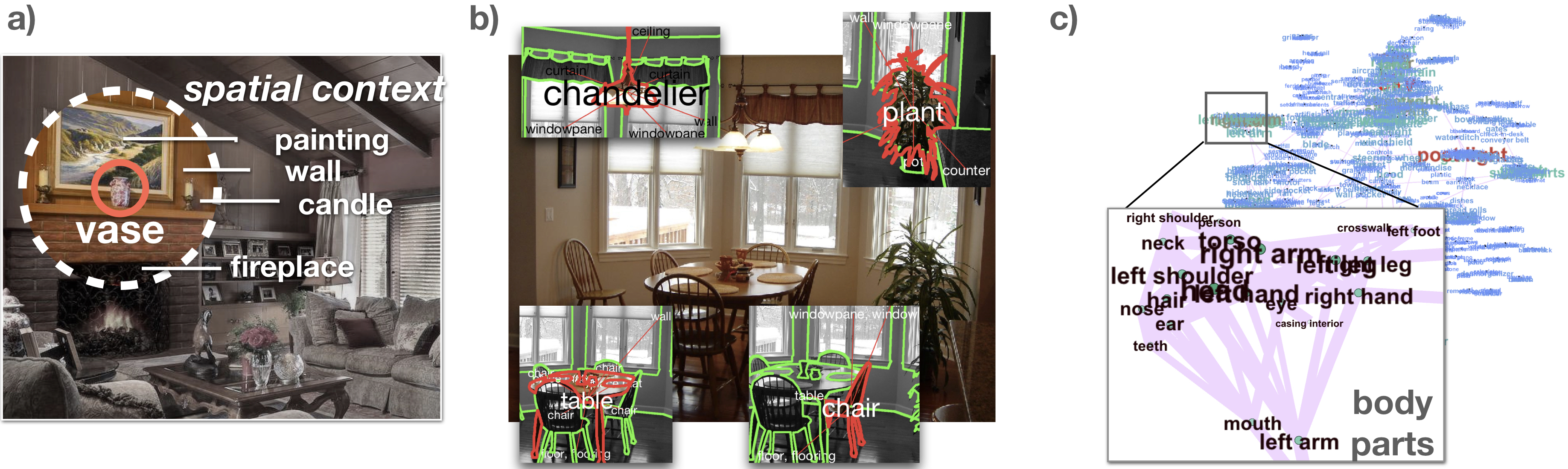}
\caption{(a) Spatial context is defined as the objects in close proximity to a target object (vase). (b) Automated analysis of spatial context for a dining room image containing 29 objects. The insets show 4 different target objects (red outline) together with their spatial neighbors (green outline). (c) Graph analysis of the object embeddings from the Skipgram model (d=300). Spatially closely related objects such as body parts appear as subclusters within the graph.}
\label{fig:spatial_context}
\end{figure*}

In the Skipgram model in Section \ref{sec:word2vec} the context of scene is defined as all objects occurring within it. While this 'bag-of-objects' approach is successful in creating meaningful scene embeddings, it is unclear whether this is an adequate approach for object embeddings. 
After all, the arrangement of objects within a scene is not random but follows semantic and syntactic rules, giving rise to hierarchies of objects within images \cite{Vo2019ReadingEnvironments}. Often, these relationships are indexed by spatial proximity. Proximity can be due to functional relatedness (a towel hanging from a towel ring or towel radiator) or object/part relationships (a body consists of different body parts). Our previous quantification of spatial relationships using co-occurrence within images probably lacks the granularity to model such tight-knit relationships. 

To alleviate this issue, we zoom in on sub-regions within images, and focus on objects and their local neighborhood. In particular, we define as \textit{spatial context} the objects in an image in close proximity to a target object (Fig. \ref{fig:spatial_context}a). Importantly, this allows us to transfer the notion of window size used by context-based methods such as Word2vec to image data. There are different ways to operationalize spatial context (e.g., radius around center, distance from object border). Furthermore, frames of reference can be proximal (image pixel coordinates) or distal (inferred 3D scene coordinates).

Here, we perform an image-based (proximal) analysis of spatial context. Since center coordinates are not always meaningful for stuff (e.g. wall) and objects with holes, we consider \textit{distance to object boundary} as a more meaningful distance metric. We parse a total of 604,355 object instances across the ADE20K dataset. For each of the images, a sparse object x object distance matrix is created with up to 345 instances per image. A nonzero entry at  position $i, j$ means that objects $i$ and $j$ are in the spatial context of each other. The calculation of the spatial context is based on the segmentation maps, where we use Matlab's \emph{imdilate} function with a 7x7 square-shaped structured element to expand a target object's boundary by 3 px. The dilated target now touches its immediate neighbors (spatial context) whose indices are obtained by intersecting all objects with the target. The distance between  target and neighbor is given by the inverse proportion of pixels the dilated target has in common with the neighbor, which implies that distances are not symmetric. For objects that share object-part relationships (e.g. hand and arm) we set the distance to a small nonzero value of $10^{-10}$. Fig. \ref{fig:spatial_context}b shows the effect of such parsing on four objects in a dining room image.

To train word embeddings we use these distance matrices with the Word2vec Skipgram model. In contrast to Section \ref{sec:word2vec}, the input to the model is now the target object and the outputs are uniformly sampled objects in the spatial context (positive examples) or outside the context (negative examples). Consequently, we maximize the probabilities

\begin{equation}\label{eq:skipgram2}
\sum_{o\in O}\sum_{o'\in\text{context}(o)}\log p(o' | o)
\end{equation}

where context(o) now refers to the local spatial context. In each iteration, 5 positive and 20 negative examples are randomly selected. For objects with less than 5 neighbors, positive examples are repeated. Hyperparameters are the same as before. The two resultant sets of embeddings are averaged. Fig. \ref{fig:spatial_context}c shows a graph analysis of the distance matrix between the embeddings thresholded at a cosine distance of 0.6. There is emergent subclusters with collections of objects that represent close spatial or object/part relationships. Another qualitative analysis is presented in Table \ref{tab:closest_neighbor}. The Skipgram model developed in this section (second column) is contrasted with the Skipgram model in Section \ref{sec:word2vec} based on object co-occurrences in scenes (third column). For the Skipgram model using spatial context, the closest objects for a given target are now indeed objects that are expected in close vicinity to the target.

\begin{table}
\footnotesize
\begin{tabular}{p{1.3cm}|p{2.6cm}|p{2.6cm}}
 \hline
 \multirow{2}{1cm}{Target object} &  \multicolumn{2}{c}{Closest neighboring objects}\\
  & Skipgram with Spatial Context & Skipgram\\
 \hline
 \multirow{3}{1cm}{towel} & towel rack 0.182& countertop 0.238\\
&towel ring 0.211 & screen door 0.259\\
&towel radiator 0.226 &shower 0.268\\
\hline
 \multirow{3}{1cm}{bicycle} & bicycle rack 0.351 & entrance 0.495\\
&parking meter 0.418 & street sign 0.501\\
&  sidewalk 0.418 & stall 0.506\\
\hline
 \multirow{3}{1cm}{pen} & notepad 0.324 & paintbrush 0.417\\
& stapler 0.427 & sandpit 0.431\\
& paper 0.430 & scotch tape 0.435\\
\hline
\end{tabular}
\caption{Closest neighbors and cosine distances for different probe objects using  Skipgram models (d=100) with spatial context. }\label{tab:closest_neighbor}
\end{table}

\section{Discussion}


Object and scene embeddings derived from object co-occurrences in images  yield semantically relevant representations. The embeddings cluster significantly along scene supercategories (indoor, urban, and natural) and subcategories (e.g., workplace, transportation, shopping and dining). This nicely dovetails with the distributional hypothesis which states that scenes are to some extent determined by the collection of objects they contain. The results are robust for a variety of models (LSA, Skipgram, CBOW) and embedding dimensions (d=50, 100, 300). Moreover, when using ADE20K's training/validation split, incorporation of LSA embeddings into a CNN for scene classification yields an improvement of 4.62\% Top5 classification accuracy over a model without object information and 3.77\% over a model that includes a vector of objects present in the scene. This suggests that the  low-dimensional representation of object vectors provided by LSA is relevant for scene classification. 

In Section \ref{sec:spatial_context}, we abandon the 'bag-of-objects' approach in order to generate object embeddings that take spatial context into account. We define spatial context as the set of objects whose boundaries touch the boundary of a target object. Embeddings based on this spatially more sophisticated approach encode meaningful and hierarchical relationships of objects that cannot be achieved with the more coarse-grained analyses.

More broadly speaking, there are two principal applications for object/scene embeddings generated with the proposed approaches. First, scene segmentation and analysis is a central problem in robotics \cite{Rangel2016SceneLabeling} which includes finding compact representations of images. In our experiment, embeddings appended to the last layer can  improve the performance of a scene classification CNN. In line with our findings, \cite{Chen2019SceneEmbeddings} show that embeddings can refine predictions of a scene classifier. These findings are corroborated by the fact that object detectors emerge in CNNs for scene classification \cite{Zhou2014ObjectCNNs}.

Second, in cognitive psychology, vision science, and sensory neuroscience scene perception, analysis, and interpretation is a model case for understanding how the brain represents complex visual information. Similar to embeddings, the human brain builds representation and expectations using image statistics and co-occurrences (both temporal and spatial) \cite{Teufel2020FormsSystem}. Since embeddings quantify object-object, scene-scene, and object-scene distances, they might provide an invaluable resource for studying the brain processes related to high-level image analysis and memory processes. For instance, object-scene pairs have been used as stimulus material in  cued memory recall experiments (e.g. \cite{Treder2020TheMemory}), and embeddings might both explain some variability observed for different object-scene combinations and guide the selection of stimulus material.

Another area, where image-based embeddings could provide an important tool, are the brain mechanisms underlying information sampling via eye-movements. For instance, it is well-established that semantic object-scene relationships have an important influence on oculomotor control in humans. Distances derived from image-based embeddings might provide a means to characterize these effects in detail, a possibility that current state-of-the-art models of oculomotor control fail to provide \cite{Pedziwiatr2019MeaningFixations}.


There are several limitations that warrant consideration. A possible critique of our first approach is that the information relating scenes to objects is already contained in the object-scene co-occurrence matrix. This is, of course,  trivially true, since the embeddings are directly built from the co-occurrences. However, we show that embeddings are able to represent this information in a lower dimensional space which is not only spatially more efficient but can also have a regularizing or denoising function  \cite{Deerwester1990IndexingAnalysis}. A related limitation is the fact that our scene classifier relies on object annotations which are not available in many datasets. However,  \cite{Chen2019SceneEmbeddings} show that such annotations can be generated on the fly  using CNNs for object detection and segmentation. Consequently, a viable approach could be to train the embeddings on annotated data and then deploy them on a system that self-generates annotations. 

Lastly, our analysis of spatial context is a proof-of-concept case, rather than a rigorous quantitative investigation. Yet, the initial findings point towards several avenues for future research. 
First, a relevant question is whether spatial context is better defined based on proximal (image-based) or distal (3D-scene based) coordinates. The latter requires inferred 3D coordinates which can be derived from image reconstruction techniques that infer depth metrics  \cite{Wu2016LearningModeling}. Second, it is conceivable that some scene categories differ not so much by the collection of objects they contain but rather by the spatial relationships between the objects. In such a case, incorporating local spatial context might be critical for good performance. Merging our approaches in sections \ref{sec:scene_embeddings} and \ref{sec:spatial_context} might be a way to address this.

In conclusion, we show that object and scene embeddings can be created from object co-occurrences and modeling of local spatial context. These embeddings are both semantically meaningful and computationally useful for downstream applications such as scene classification.



\bibliography{references}

\bibliographystyle{abbrv}

\end{document}